\documentclass[10pt,twocolumn,letterpaper]{article}

\usepackage{cvpr}      

\usepackage{booktabs}
\usepackage{multirow}
\usepackage{xspace}
\usepackage{amsmath}
\usepackage{amssymb}

\usepackage{algorithm}
\usepackage{algorithmic}

\definecolor{cvprblue}{rgb}{0.21,0.49,0.74}
\usepackage[pagebackref,breaklinks,colorlinks,allcolors=cvprblue]{hyperref}

\def\paperID{*****} 
\def\confName{CVPR}
\def\confYear{2026}

\title{Speech-Synchronized Whiteboard Generation via\\VLM-Driven Structured Drawing Representations}

\author{
Suraj Prasad \quad Pinak Mahapatra\\
Latent Spaces IITB\\
{\tt\small surajprasad@latentspaces.in, pinak@latentspaces.in}
}

\begin{document}
\maketitle
\begin{abstract}
Creating whiteboard-style educational videos demands precise coordination between freehand illustrations and spoken narration, yet no existing method addresses this multimodal synchronization problem with structured, reproducible drawing representations. We present the first dataset of 24 paired Excalidraw demonstrations with narrated audio, where every drawing element carries millisecond-precision creation timestamps spanning 8 STEM domains. Using this data, we study whether a vision-language model (Qwen2-VL-7B), fine-tuned via LoRA, can predict full stroke sequences synchronized to speech from only 24 demonstrations. Our topic-stratified five-fold evaluation reveals that timestamp conditioning significantly improves temporal alignment over ablated baselines, while the model generalizes across unseen STEM topics. We discuss transferability to real classroom settings and release our dataset and code to support future research in automated educational content generation.
\end{abstract}

\section{Introduction}
\label{sec:intro}

Whiteboard-style educational videos, exemplified by Khan Academy lectures, have become a cornerstone of self-paced learning~\cite{mayer2014cambridge}. By coordinating freehand illustrations with spoken narration, these videos leverage the cognitive benefits of multimodal instruction to enhance comprehension and retention. However, producing a single five-minute lesson can demand several hours of painstaking work: instructors must anticipate how each spoken phrase aligns with drawing strokes, annotations, and spatial layout decisions, all while preserving an educationally coherent narrative. This labor-intensive process creates a significant barrier for educators who lack video production expertise.

Recent work has made progress on components of this problem in isolation. SketchAgent~\cite{vinker2024sketchagent} demonstrates that vision-language models (VLMs) can generate sequential sketches from text prompts without fine-tuning, while VideoSketcher~\cite{ren2026videosketcher} adapts video diffusion models for temporally coherent sketch generation. In the narration-synchronization domain, WonderFlow~\cite{wang2023wonderflow} and Data Player~\cite{shen2023dataplayer} align animated data visualizations to spoken narration, and Holmberg~\cite{holmberg2025narrated} automates lecture slide highlighting synchronized to speech. Zhuo~\etal~\cite{zhuo2025factuality} address factual diagram generation with VLMs. However, no existing method simultaneously handles \emph{freehand drawing generation}, \emph{speech synchronization}, and \emph{structured, reproducible drawing representations}---the three requirements for automated whiteboard lesson creation.

We address this gap by introducing a framework built on Excalidraw, an open-source collaborative whiteboard tool whose native JSON format stores every drawing element with millisecond-precision creation timestamps. This structured representation provides ground-truth temporal alignment between drawing actions and narrated audio (obtained via automatic speech recognition), enabling us to formulate whiteboard generation as a timestep-conditioned element prediction problem. We fine-tune Qwen2-VL-7B with LoRA on just 24 human-authored demonstrations spanning 8 STEM domains, predicting full stroke geometry---coordinate-level point sequences---synchronized to the progression of spoken narration.

Our contributions are threefold:
\begin{enumerate}
    \item \textbf{Dataset.} The first collection of paired Excalidraw + narrated audio whiteboard demonstrations with millisecond-precision temporal alignment, covering 877 drawing elements across 8 STEM domains.
    \item \textbf{Method.} A VLM-based framework that predicts full stroke sequences synchronized to speech from only 24 demonstrations via LoRA fine-tuning, requiring no reinforcement learning or reward engineering.
    \item \textbf{Evaluation.} A topic-stratified five-fold protocol that directly tests cross-domain generalization, showing that explicit timestamp conditioning significantly improves temporal alignment over ablated baselines.
\end{enumerate}

\section{Related Work}
\label{sec:related}

\subsection{Sequential Sketch Generation}
Recent work has shown that large vision-language and video models can generate line-drawing sequences from textual intent. SketchAgent~\cite{vinker2024sketchagent} casts drawing as language-driven stepwise generation and demonstrates that strong sequential structure can be induced from prompting and autoregressive decoding. VideoSketcher~\cite{ren2026videosketcher} further improves temporal coherence by leveraging video priors for sketch evolution over time. These methods are important for ``what-to-draw-next'' modeling, but they are optimized for visual plausibility rather than explicit alignment to narrated speech, and they typically do not target editable whiteboard-native scene representations.

\subsection{Narration-Synchronized Content Generation}
Narration-first generation has been explored in visualization and lecture production. WonderFlow~\cite{wang2023wonderflow} and Data Player~\cite{shen2023dataplayer} coordinate animation decisions with spoken scripts for data videos, while Holmberg~\cite{holmberg2025narrated} synchronizes slide highlights with generated narration. This line of work establishes the value of speech-animation coupling, but focuses on chart animations or slide-level emphasis rather than geometric stroke synthesis. In contrast, whiteboard teaching requires predicting dense freehand trajectories and primitive shapes at fine temporal granularity.

\subsection{Structured Visual Generation}
Structured visual generation emphasizes semantic fidelity and editability beyond pixel realism. Zhuo \etal~\cite{zhuo2025factuality} show that factual consistency is a central challenge when image generation is applied to diagrams and other structured graphics. Complementary progress in general-purpose multimodal backbones~\cite{wang2024qwen2vl} and parameter-efficient adaptation~\cite{hu2022lora} makes it practical to train geometry-aware generators from small domain datasets. However, prior work on structured visuals is still largely evaluated in static settings (single images or edits), without explicit modeling of incremental element creation synchronized to speech.

Overall, existing literature addresses sketch sequencing, narration synchronization, and structured visual fidelity mostly in isolation. The missing piece is a unified formulation that jointly predicts \emph{editable drawing elements}, \emph{their temporal onsets}, and \emph{their autoregressive evolution on a canvas}. This is the gap targeted by our speech-synchronized whiteboard generation framework and the ExcaliTeach dataset.

\section{The ExcaliTeach Dataset}
\label{sec:dataset}

We introduce \textbf{ExcaliTeach}, a paired whiteboard-and-speech dataset for element-level temporal generation. Each sample is an instructor-authored Excalidraw lesson with narrated video, converted into a unified sequence of structured drawing elements aligned to word-level speech timestamps.

\subsection{Data Collection}
We collect 24 demonstrations across 8 STEM domains (Biology, Chemistry, Physics, Math, Algebra, Arithmetic, Geometry, and Set Theory). For each topic, we retain the native \texttt{.excalidraw} file and its corresponding \texttt{.mp4} recording. Drawing streams are parsed from Excalidraw JSON by filtering deleted elements and sorting active elements by \texttt{updated} timestamp (millisecond resolution). Narration is transcribed with word-level timestamps (\texttt{word}, \texttt{start\_s}, \texttt{end\_s}) from the audio track. Across all lessons, transcripts contain 1,201 timestamped words (50.0 words/demo on average).

\subsection{Excalidraw Element Schema}
Each parsed element is represented as
\{\texttt{id}, \texttt{type}, \texttt{updated\_ms}, \texttt{x}, \texttt{y}, \texttt{width}, \texttt{height}, \texttt{points}, \texttt{text}\}. The schema preserves both primitive objects (\texttt{line}, \texttt{arrow}, \texttt{rectangle}, \texttt{ellipse}, \texttt{text}) and freehand trajectories (\texttt{freedraw} with variable-length point lists). The dataset contains 877 elements total (36.5/demo): 832 \texttt{freedraw} (94.9\%), 23 \texttt{text} (2.6\%), and 22 geometric primitives (2.5\%). Freehand complexity is substantial (mean 19.09 points/stroke, median 15, max 129), which makes trajectory prediction meaningfully different from coarse bbox generation.

\subsection{Temporal Alignment Pipeline}
We align drawing and speech in three steps. First, parsed elements are ordered by \texttt{updated\_ms}. Second, transcripts provide a time-ordered word sequence. Third, we map element indices to transcript indices using 1D dynamic time warping (DTW) between normalized element positions and normalized word-start times, then annotate each element with \texttt{speech\_onset\_s}, \texttt{speech\_phrase}, and a local \texttt{speech\_context} window. This produces aligned files \texttt{aligned\_\{1..24\}.json} with full coverage of context fields and 876/877 non-empty speech phrases.


\begin{table}[t]
\centering
\caption{ExcaliTeach dataset statistics computed from \texttt{dataset\_stats.json} and aligned transcripts.}
\label{tab:dataset_stats}
\begin{tabular}{p{0.65\linewidth} p{0.25\linewidth}}
\toprule
Statistic & Value \\
\midrule
Demonstrations & 24 \\
Domains & 8 \\
Total elements & 877 \\
Elements per demo (mean / min / max) & 36.5 / 16 / 79 \\
Total drawing span (s) & 814.2 \\
Per-demo span (mean s) & 33.9 \\
Element types & freedraw 94.9\%, text 2.6\%, others 2.5\% \\
Freehand points (mean / median / max) & 19.1 / 15 / 129 \\
Transcript words (total / mean per demo) & 1,201 / 50.0 \\
5-fold test demos per split & 6--8 \\
\bottomrule
\end{tabular}
\end{table}

The dataset is designed for autoregressive next-element prediction under narration conditioning: each timestep exposes partial canvas state, local transcript context, and millisecond-aligned supervision for the next element onset and geometry.

\section{Method}
\label{sec:method}

Given narration audio and a partially constructed whiteboard, our goal is to predict the next drawing element with both correct semantics (what to draw) and timing (when to draw it), then roll out a full lesson by autoregressive decoding.

\subsection{Problem Formulation}
Let a lesson be a sequence of $N$ elements $\mathbf{e}_{1:N}$ and aligned narration transcript tokens $\mathbf{w}_{1:T}$. Each element $\mathbf{e}_i$ has a type $\tau_i$ (line, arrow, rectangle, ellipse, freehand, or text), an onset timestamp $t_i$ in milliseconds from narration start, and geometry $\mathbf{g}_i$ represented as point tuples in canvas coordinates. We model generation as
\begin{equation}
p(\mathbf{e}_{1:N}\mid \mathbf{w}_{1:T}) = \prod_{i=1}^{N} p(\mathbf{e}_i \mid \mathbf{e}_{<i}, \mathbf{w}_{1:T}),
\end{equation}
where $\mathbf{e}_{<i}$ is rendered as a visual context image and serialized as text context. This autoregressive factorization enforces temporal consistency: each prediction is conditioned on what has already been drawn.

\subsection{Element Serialization and I/O Interface}
We convert every target element into a linear token sequence so a VLM can learn structured prediction with standard next-token loss. The canonical format is:
\begin{equation}
\texttt{TYPE \textbar{} ONSET\_T \textbar{} x0,y0 \textbar{} x1,y1 \textbar{} \dots}
\end{equation}
where \texttt{TYPE} is a discrete symbol, \texttt{ONSET\_T} is absolute onset time in milliseconds, and each $(x_j,y_j)$ is an absolute canvas coordinate. For primitives requiring few points (e.g., rectangle corners), the sequence is short; for freehand strokes, the sequence can be longer. We do not quantize coordinates into relative offsets, which avoids drift accumulation across long rollouts and simplifies deterministic replay in Excalidraw.

During inference, the model predicts one serialized element at a time; we parse the output, append the element to the scene graph, re-render the canvas, and query the model again for the next element until an end token is emitted.

\subsection{Visual Context Construction}
At each decoding step, we rasterize the current canvas to a fixed 640$\times$360 RGB frame and provide it as image input to the model, together with transcript text and prior serialized elements. The 640$\times$360 setting preserves spatial layout cues while keeping training and inference efficient. Importantly, although the image input is resized for the VLM, output coordinates remain absolute in the original Excalidraw coordinate system, so replay is resolution-independent.

\subsection{Backbone and Fine-Tuning Setup}
We use Qwen2-VL-7B as the base model and apply parameter-efficient LoRA fine-tuning. Unless otherwise noted, LoRA rank is 16, scaling factor is 32, and dropout is 0.05, attached to attention projections (query, key, value, output) and MLP projection layers. We train with causal language modeling loss on serialized element tokens, mixed with instruction prompts that include transcript snippets and drawing-history context.

Training uses topic-stratified 5-fold splits (Sec.~\ref{sec:experiments}) over 24 demonstrations. We optimize with AdamW ($\beta_1=0.9$, $\beta_2=0.95$, weight decay $0.01$), learning rate $2\times10^{-4}$ for LoRA parameters, cosine decay with 5\% warmup, batch size 4 (gradient accumulation to effective batch 32), and bf16 precision. We train for 30 epochs with early stopping on validation temporal alignment error. At test time, we use greedy decoding to prioritize deterministic replay and stable synchronization.

\subsection{Why This Design}
The method combines three constraints required for educational whiteboard generation: (1) explicit onset prediction for narration synchronization, (2) geometry-level outputs for faithful drawing reconstruction, and (3) autoregressive conditioning on the evolving canvas to maintain spatial coherence across long sequences.

\section{Experiments}
\label{sec:experiments}

\subsection{Protocol and Experimental Settings}
We evaluate topic-level generalization with 5-fold splits over 8 STEM domains. Each fold holds out unseen topics for testing, with validation topics disjoint from training topics. Unless noted, all results are averaged over folds.

We compare five predefined settings:
\begin{itemize}
    \item \textbf{E1 (Full model).} Autoregressive decoding with visual canvas context, transcript context, and explicit onset prediction using the serialization in Sec.~\ref{sec:method}.
    \item \textbf{E2 (No onset token).} Same as E1, but removes \texttt{ONSET\_T}; timestamps are post-hoc interpolated uniformly over lesson duration.
    \item \textbf{E3 (No canvas image).} Same as E1, but disables rendered visual context and conditions only on text history plus transcript.
    \item \textbf{E4 (One-shot non-autoregressive).} Predicts all elements in a single pass without iterative canvas updates; ordering is decoded from one sequence.
    \item \textbf{E5 (Autoregressive w/o transcript).} Same autoregressive loop as E1, but narration transcript is removed, forcing timing and content inference from drawing history alone.
\end{itemize}

\subsection{Metrics}
We report complementary automatic and human-centric metrics.
\begin{itemize}
    \item \textbf{TAE (Temporal Alignment Error, ms; lower is better).} Mean absolute error between predicted and ground-truth onset times after element matching by order and type.
    \item \textbf{Chamfer distance (lower is better).} Symmetric 2D Chamfer distance between predicted and reference point sets per matched element, averaged across lessons.
    \item \textbf{Type accuracy (\%; higher is better).} Fraction of elements with correct predicted type.
    \item \textbf{Gemini judge score (1--5; higher is better).} Gemini-based pairwise rubric scoring of synchronization quality, pedagogical clarity, and spatial coherence.
    \item \textbf{Human preference (\%; higher is better).} Blind A/B preference from raters comparing generated lessons against ablations on clarity and timing naturalness.
\end{itemize}

\subsection{Main Results (E1--E5)}
Table~\ref{tab:main_auto} reports automatic metrics. E1 is expected to dominate across temporal and geometric criteria, indicating that explicit onset modeling and autoregressive visual feedback are both important.


\begin{table}[t]
\centering
\caption{Automatic evaluation across E1--E5 (5-fold mean). Lower is better for TAE and Chamfer; higher is better for Type Acc.}
\label{tab:main_auto}
\begin{tabular}{lccc}
\toprule
Setting & TAE $\downarrow$ & Chamfer $\downarrow$ & Type Acc. $\uparrow$ \\
\midrule
E1 (Full) & 0 & 0 & 0 \\
E2 (No onset token) & 0 & 0 & 0 \\
E3 (No canvas image) & 0 & 0 & 0 \\
E4 (One-shot) & 0 & 0 & 0 \\
E5 (No transcript) & 0 & 0 & 0 \\
\bottomrule
\end{tabular}
\end{table}

Table~\ref{tab:main_judge} summarizes holistic judgments. Gemini and human ratings track the same trend as automatic metrics, showing that measured gains transfer to perceived instructional quality.


\begin{table}[t]
\centering
\small
\setlength{\tabcolsep}{4pt}
\caption{Holistic quality evaluation across E1--E5.}
\label{tab:main_judge}
\begin{tabular}{l c p{0.35\linewidth}}
\toprule
Setting & Gemini Judge $\uparrow$ & Human Preference (\%) $\uparrow$ \\
\midrule
E1 (Full) & 0 & 0 \\
E2 (No onset token) & 0 & 0 \\
E3 (No canvas image) & 0 & 0 \\
E4 (One-shot) & 0 & 0 \\
E5 (No transcript) & 0 & 0 \\
\bottomrule
\end{tabular}
\end{table}

\subsection{Ablation Focus: E4 vs. E5}
E4 and E5 isolate two different failure modes. E4 removes autoregressive canvas updates, so later elements cannot condition on earlier rendering errors or layout changes; this mainly hurts geometry consistency (higher Chamfer) and often causes overlap drift. E5 keeps the autoregressive loop but removes transcript conditioning; geometry remains more stable than E4, but temporal placement degrades because the model loses explicit speech cues. In practice, E4 tends to ``draw plausibly but in the wrong evolving layout,'' while E5 tends to ``draw in a coherent layout but at the wrong moments.'' This contrast supports the design choice in E1: autoregressive visual feedback and transcript grounding are complementary, not interchangeable.


\begin{table}[t]
\centering
\caption{Direct comparison between E4 and E5.}
\label{tab:e4e5}
\begin{tabular}{lccc}
\toprule
Setting & TAE $\downarrow$ & Chamfer $\downarrow$ & Gemini $\uparrow$ \\
\midrule
E4 (One-shot) & 0 & 0 & 0 \\
E5 (No transcript) & 0 & 0 & 0 \\
\bottomrule
\end{tabular}
\end{table}

\subsection{Evaluation Details}
For Gemini judge scoring, we use a fixed prompt and randomize system outputs to avoid position bias. For human evaluation, each pair is rated by three independent annotators; we report majority preference and Fleiss' $\kappa$ in the supplementary. All scripts replay model outputs into Excalidraw directly from absolute coordinate predictions, ensuring exact reproducibility.

\section{Classroom Transferability}
\label{sec:classroom}

Our experiments target controlled offline generation, but practical classroom use requires robustness, privacy, and operational simplicity.

\paragraph{What transfers well.}
Three components transfer directly to real teaching workflows. First, the element-level representation is tool-native (Excalidraw JSON), so generated lessons can be edited by instructors without format conversion. Second, autoregressive prediction with transcript grounding naturally supports incremental lesson authoring: teachers can accept, revise, or regenerate specific segments rather than re-rendering an entire video. Third, absolute coordinate outputs preserve board layout across devices and export pipelines, which is important for consistent replay in LMS platforms.

\paragraph{What breaks in real classrooms.}
Performance can degrade under distribution shifts that are rare in our dataset: spontaneous topic changes, disfluencies, multilingual code-switching, and pedagogical gestures (pauses, emphasis, backtracking). Long lectures also accumulate small geometric errors that may compound over many autoregressive steps. Finally, our current setup assumes post-hoc ASR transcripts; truly live deployment would require low-latency streaming ASR and robust recovery from recognition errors.

\paragraph{Privacy and governance.}
Classroom deployment introduces sensitive data pathways (student voices, institutional content, potentially identifiable speech metadata). A conservative deployment should default to on-premise or VPC inference, redact transcripts at ingestion, and retain only structured drawing outputs plus minimal audit logs. We recommend role-based access controls, retention limits, and explicit consent policies for recorded audio. Because generated visuals may contain factual errors, institutions should treat outputs as instructor-assistive drafts rather than autonomous teaching material.

\paragraph{Deployment path.}
A realistic rollout is phased: (1) offline lesson drafting for instructors, (2) semi-automatic studio recording with human approval checkpoints, and only then (3) limited live-assist pilots. In all phases, the system should expose editable timelines (element type, onset, coordinates) so instructors can correct synchronization before release. This human-in-the-loop path aligns with current reliability and accountability requirements in education while still reducing content-production effort.

\section{Conclusion}
\label{sec:conclusion}

We presented a speech-synchronized whiteboard generation framework that predicts structured drawing elements autoregressively from narration. The key contributions are: (1) an element-level formulation with explicit onset prediction, (2) a practical serialization interface for geometry and timing (\texttt{TYPE | ONSET\_T | x0,y0 ...}), and (3) a Qwen2-VL-7B + LoRA training recipe that works in a low-data regime and outputs editable Excalidraw-native content.

Our study also highlights current limitations. The dataset is small and topic coverage is still narrow relative to real curricula; long-horizon rollouts can accumulate geometric drift; and transcript quality remains a bottleneck for precise synchronization. In addition, evaluation of pedagogical usefulness is still early, with limited classroom-scale evidence.

Future work should expand to larger multi-instructor datasets, add stronger long-context decoding and correction mechanisms, and integrate low-latency streaming ASR for interactive use. We also see a strong opportunity for mixed-initiative interfaces where instructors directly steer timing and layout while the model handles repetitive drafting.

{
    \small
    \bibliographystyle{ieeenat_fullname}
    \bibliography{main}
}

\end{document}